# Estimating the Success of Unsupervised Image to Image Translation

**Sagie Benaim**[*]                                    SAGIEB@MAIL.TAU.AC.IL
**Tomer Galanti**[*]                                   TOMERGA2@POST.TAU.AC.IL
*The Blavatnik School of Computer Science*
*Tel Aviv University*
*Tel Aviv, Israel*

**Lior Wolf**                                          WOLF@FB.COM
*Facebook AI Research &*
*The Blavatnik School of Computer Science*
*Tel Aviv University*
*Tel Aviv, Israel*

## Abstract

While in supervised learning, the validation error is an unbiased estimator of the generalization (test) error and complexity-based generalization bounds are abundant, no such bounds exist for learning a mapping in an unsupervised way. As a result, when training GANs and specifically when using GANs for learning to map between domains in a completely unsupervised way, one is forced to select the hyperparameters and the stopping epoch by subjectively examining multiple options. We propose a novel bound for predicting the success of unsupervised cross domain mapping methods, which is motivated by the recently proposed Simplicity Principle. The bound can be applied both in expectation, for comparing hyperparameters and for selecting a stopping criterion, or per sample, in order to predict the success of a specific cross-domain translation. The utility of the bound is demonstrated in an extensive set of experiments employing multiple recent algorithms. Our code is available at `https://github.com/sagiebenaim/gan_bound`.

## 1. Introduction

In unsupervised learning, the process of selecting hyperparameters and the lack of clear stopping criteria are a constant source of frustration. This issue is commonplace for GANs [1] and the derived technologies, in which the training process optimizes multiple losses that balance each other. Practitioners are often uncertain regarding the results obtained when evaluating GAN-based methods, and many avoid using these altogether. One solution is to employ more stable methods such as [2]. However, these methods do not always match the results obtained by GANs. In this work, we offer, for an important family of GAN methodologies, an algorithm for selecting the hyperparameters, as well as a stopping criterion.

---

[*]. Authors contributed equally

Specifically, we focus on predicting the success of algorithms that map between two image domains in an unsupervised manner. Multiple GAN-based methods have recently demonstrated convincing results, despite the apparent inherent ambiguity, which is described in Sec. 2. We derive what is, as far as we know, the first error bound for unsupervised cross domain mapping.

In addition to the novel capability of predicting the success, in expectation, of a mapping that was trained using one of the unsupervised mapping methods, we can predict the success of mapping every single sample individually. This is remarkable for two reasons: (i) even supervised generalization bounds do not deliver this capability; and (ii) we deal with complex multivariate regression problems (mapping between images) and not with classification problems, in which pseudo probabilities are often assigned.

In Sec. 2, we formulate the problem and present background on the Simplicity Principle of [3]. Then, in Sec. 3, we derive the prediction bounds and introduce multiple algorithms. Sec. 4 presents extensive empirical evidence for the success of our algorithms, when applied to multiple recent methods. This includes a unique combination of the hyperband method [4], which is perhaps the leading method in hyperparameter optimization, in the supervised setting, with our bound. This combination enables the application of hyperband in unsupervised learning, where, as far as we know, no hyperparameter selection method exists.

## 1.1 Related Work

**Generative Adversarial Networks** GAN [1] methods train a generator network $G$ that synthesizes samples from a target distribution, given noise vectors, by jointly training a second, adversarial, network $D$. Conditional GANs employ a vector of parameters that directs the generator, in addition to (or instead of) the noise vector. These GANs can generate images from a specific class [5] or based on a textual description [6], or invert mid-level network activations [7]. Our bound also applies in these situations. However, this is not the focus of our experiments, which target image mapping, in which the created image is based on an input image [8, 9, 10, 11, 12, 13, 14].

**Unsupervised Mapping** The validation of our bound focuses on recent cross-domain mapping methods that employ no supervision, except for sample images from the two domains. This ability was demonstrated recently [8, 9, 10, 14] in image to image translation and slightly earlier for translating between natural languages [15].

The DiscoGAN [8] method, similar to other methods [9, 10], learns mappings in both directions, i.e., from domain $A$ to domain $B$ and vice versa. Our experiments also employ the DistanceGAN method [14], which unlike the circularity based methods, is applied only in one direction (from $A$ to $B$). The constraint used by this method is that the distances for a pair of inputs $x_1, x_2 \in A$ before and after the mapping, by the learned mapping $G$, are highly correlated, i.e., $||x_1 - x_2|| \sim ||G(x_1) - G(x_2)||$.

**Weakly Supervised Mapping** Our bound can also be applied to GAN-based methods that match between the source domain and the target domain by also incorporating a fixed pre-trained feature map $f$ and requiring $f$-constancy, i.e, that the activations of $f$ are the same for the input samples and for mapped samples [12, 16]. During training, the various components of the loss (GAN, f-constancy, and a few others) do not provide a clear signal when to stop training or which hyperparameters to use.

**Generalization Bounds for Unsupervised Learning** Only a few generalization bounds for unsupervised learning were suggested in the literature. In [17], PAC-Bayesian generalization bounds are presented for density estimation. [18] gives an algorithm for estimating a bounded density using a finite combination of densities from a given class. This algorithm has estimation error bounded by $O(1/\sqrt{n})$. Our work studies the error of a mapping and not the KL-divergence with respect to a target distribution. Further, our bound is data-dependent and not based on the complexity of the hypothesis class.

**Hyperparameter Optimization** Hyperparameters are constants and configurations that are being used by a learning algorithm. Hyperparameter selection is the process of selecting the hyperparameters that will produce better learning. This includes optimizing the number of epochs, size and depth of the neural network being trained, learning rate, etc. Many of the earlier hyperparameter methods that go beyond a random- or a grid-search were Bayesian in nature [19, 20, 21, 22, 23]. The hyperband method [4], which is currently leading various supervised learning benchmarks, is based on the multi-arm bandit problem. It employs partial training and dynamically allocates more resources to successful configurations. All such methods crucially rely on a validation error to be available for a given configuration, which means that these can only be used in the supervised settings. Our work enables, for the first time, the usage of such methods also in the unsupervised setting, by using our bound in lieu of the validation error for predicting the ground truth error.

## 2. Problem Setup

In Sec. 2.1 we define the alignment problem. Sec 2.2 illustrates the Simplicity Principle which was introduced in [3] and was verified with an extensive set of experiments. Sec. 2.3 and everything that follows are completely novel. The section proposes the Occam's razor property, which extends the definition of the Simplicity Principle, and which is used in Sec. 3 to derive the main results and algorithms.

### 2.1 The Alignment Problem

The learning algorithm is provided with two unlabeled datasets: one includes i.i.d samples from a first distribution and the second, i.i.d samples from a second distri-bution.

$$S_A = \{x_i\}_{i=1}^m \stackrel{\text{i.i.d}}{\sim} D_A^m \text{ and } S_B = \{y_i\}_{i=1}^n \stackrel{\text{i.i.d}}{\sim} D_B^n \quad (1)$$

$D_A$ and $D_B$ are distributions over $\mathcal{X}_A$ and $\mathcal{X}_B$ (resp.). In addition, $y_{AB}$ denotes the target function, which is one of the functions that map the first domain to the second, such that $y_{AB} \circ D_A = D_B$ ($g \circ D$ is defined to be the distribution of $g(x)$ where $x \sim D$). The goal of the learner is to fit a function $G \in \mathcal{H}$, for some hypothesis class $\mathcal{H}$ that is closest to $y_{AB}$, i.e,

$$\inf_{G \in \mathcal{H}} R_{D_A}[G, y_{AB}] \quad (2)$$

where $R_D[f_1, f_2] = \mathbb{E}_{x \sim D}[\ell(f_1(x), f_2(x))]$, for a loss function $\ell: \mathbb{R}^M \times \mathbb{R}^M \to \mathbb{R}$ and distribution $D$.

It is not clear that such fitting is possible, without additional information. Assume, for example, that there is a natural order on the samples in $\mathcal{X}_B$. A mapping that maps an input sample $x \in \mathcal{X}_A$ to the sample that is next in order to $y_{AB}(x)$, could be just as feasible. More generally, one can permute the samples in $\mathcal{X}_A$ by some function $\Pi$ that replaces each sample with another sample that has a similar likelihood and learn $G$ that satisfies $G = \Pi \circ y_{AB}$. This difficulty is referred to in [3] as "the alignment problem".

In multiple recent contributions [15, 8, 9, 10], circularity is employed. Circularity requires the recovery of both $y_{AB}$ and $y_{BA} = y_{AB}^{-1}$ simultaneously. Namely, functions $G$ and $G'$ are learned jointly by minimizing the following objective:

$$\text{disc}(G \circ D_A, D_B) + \text{disc}(G' \circ D_B, D_A) + R_{D_A}[G' \circ G, \text{Id}_A] + R_{D_B}[G \circ G', \text{Id}_B] \quad (3)$$

where

$$\text{disc}(D_1, D_2) = \sup_{c_1, c_2 \in \mathcal{C}} \left| R_{D_1}[c_1, c_2] - R_{D_2}[c_1, c_2] \right| \quad (4)$$

denotes the discrepancy between distributions $D_1$ and $D_2$, and $\mathcal{C}$ is a set of discriminators. This discrepancy is implemented by a GAN, as in [24].

As shown in [3], the circularity constraint does not eliminate the uncertainty in its entirety. In DistanceGAN [14], the circularity was replaced by a multidimensional scaling type of constraint, which enforces a high correlation between the distances in the two domains. However, since these constraints hold only approximately, the ambiguity is not completely eliminated.

## 2.2 The Simplicity Principle

In order to understand how the recent unsupervised image mapping methods work despite the inherent ambiguity, [3] recently showed that the target ("semantic") mapping $y_{AB}$ is typically the distribution preserving mapping ($h \circ D_A = D_B$) with the lowest complexity. It was shown that such mappings are expected to be unique.

As a motivating example to the key role of minimal mappings, consider the domain $A$ of uniformly distributed points $(x_1, x_2)^\top \in \mathbb{R}^2$, where $x_1 = x_2 \in [-1, 1]$. Let $B$ be the domain of uniformly distributed points in $\{(x_1, x_2)^\top | x_1 \in [0, 1], x_2 = 0\} \cup$ $\{(x_1, x_2)^\top | x_2 \in [0,1], x_1 = 0\}$. We note that there are infinitely many mappings from domain $A$ to $B$ that, given inputs in $A$, result in the uniform distribution of $B$ and satisfy the circularity constraint (Eq. 3).

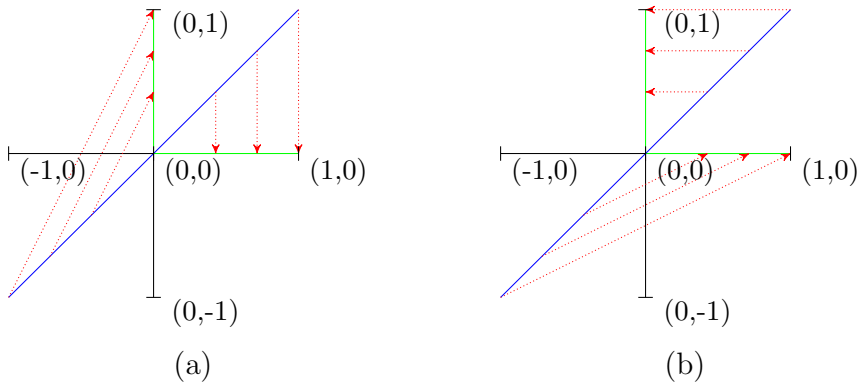

Figure 1: An illustrative example, where the two domains are the blue and green areas. There are infinitely many mappings that preserve the uniform distribution on the two domains. However, only two stand out as "semantic". These two, which are depicted in red, are exactly the two mappings that can be captured by a minimal neural network with ReLU activations. (a) the mapping $y^1_{AB}$. (b) the mapping $y^2_{AB}$ (see Eq. 5).

However, it is easy to see that when restricting the hypothesis class to neural networks with one layer of size 2, and ReLU activations $\sigma$, there are only two options left. In this case, $h(x) = \sigma_a(Wx)$, for $W \in \mathbb{R}^{2\times 2}, b \in \mathbb{R}^2$. The only admissible solutions are of the form $W = \begin{pmatrix} a & 1-a \\ b & -1-b \end{pmatrix}$ or $W' = \begin{pmatrix} a & -1-a \\ b & 1-b \end{pmatrix}$, which are identical, for every $a, b \in \mathbb{R}$, to one of the following functions:

$$y^1_{AB}((x,x)^\top) = \begin{cases} (x,0)^\top & \text{if } x \geq 0 \\ (0,-x)^\top & \text{if } x \leq 0 \end{cases} \text{ and } y^2_{AB}((x,x)^\top) = \begin{cases} (0,x)^\top & \text{if } x \geq 0 \\ (-x,0)^\top & \text{if } x \leq 0 \end{cases} \quad (5)$$

Therefore, by restricting the hypothesis space to be minimal, we eliminate all alternative solutions, except two. These two are exactly the two mappings that would commonly be considered "more semantic" than any other mapping, see Fig. 1. Another motivating example can be found in [3].

### 2.3 Occam's Razor

We note that the Simplicity Principle, presented in [3], is highly related to the principle known as Occam's razor. In this section we provide a definition of the Occam's razor property which extends the formulation of the Simplicity Principle used in [3]. Our formulation is not limited to Kolmogorov-like complexity of multi-layered neural networks as in [3] and is more general.

Given two domains $A = (\mathcal{X}_A, D_A)$ and $B = (\mathcal{X}_B, D_B)$, a mapping $y_{AB} : \mathcal{X}_A \to \mathcal{X}_B$ satisfies the Occam's razor property between domains $A$ and $B$, if it has minimal complexity among the functions $h : \mathcal{X}_A \to \mathcal{X}_B$ that satisfy $h \circ D_A \approx D_B$. Minimal complexity is defined by the nesting of hypothesis classes, which forms a partial order, and not as a continuous score. For example, if $\mathcal{H}_j$ is the set of neural networks of a specific architecture and $\mathcal{H}_i$ is the set of neural networks of the architecture obtained after deleting one of the hidden neurons, then, $\mathcal{H}_i \subset \mathcal{H}_j$. Intuitively, minimal complexity would mean that there is no sub-class that can implement a mapping $h : \mathcal{X}_A \to \mathcal{X}_B$ such that $h \circ D_A \approx D_B$. For this purpose, we define,

$$\mathcal{P}(\mathcal{H}; \epsilon) = \left\{ G \in \mathcal{H} \,\middle|\, \text{disc}(G \circ D_A, D_B) \leq \epsilon \right\} \tag{6}$$

**Definition 1** (Occam's razor property). *Let $A = (\mathcal{X}_A, D_A)$ and $B = (\mathcal{X}_B, D_B)$ be two domains and $\mathcal{U} = \{\mathcal{H}_i\}_{i \in I}$ be a family of hypothesis classes. A mapping $y_{AB} : \mathcal{X}_A \to \mathcal{X}_B$ satisfies an $(\epsilon_1, \epsilon_2)$-Occam's razor property if for every $\mathcal{H} \in \mathcal{U}$ such that $\mathcal{P}(\mathcal{H}; \epsilon_1) \neq \emptyset$, we have:* $\inf_{G \in \mathcal{P}(\mathcal{H}; \epsilon_1)} R_{D_A}[G, y_{AB}] \leq \epsilon_2$.

Informally, according to Def. 1, a function satisfies the Occam's razor property, if it can be approximated by even the lowest-complexity hypothesis classes that successfully map between the domains $A$ and $B$. If $y_{AB}$ has the $(\epsilon_1, \epsilon_2)$-Occam's razor property, then it is $\epsilon_2$-close to a function in every minimal hypothesis class $\mathcal{H} \in \mathcal{U}$ such that $\mathcal{P}(\mathcal{H}; \epsilon_1) \neq \emptyset$. As the hypothesis class $\mathcal{H}$ grows, so does $\mathcal{P}(\mathcal{H}; \epsilon_1)$, i.e., $\mathcal{H}_i \subset \mathcal{H}_j$ implies that $\mathcal{P}(\mathcal{H}_i; \epsilon_1) \subset \mathcal{P}(\mathcal{H}_j; \epsilon_1)$. Therefore, the growing $\mathcal{P}(\mathcal{H}; \epsilon_1)$ would always contain at least one function that is $\epsilon_2$-close to $y_{AB}$. Nevertheless, as the hypothesis class grows, $\mathcal{P}(\mathcal{H}; \epsilon_1)$ can potentially contain many functions $f$ that satisfy $f \circ D_A \approx D_B$ and differ from each other, causing an increased amount of ambiguity. In addition, we note that uniqueness is not assumed, and the property may hold for multiple mappings.

## 3. Estimating the Ground Truth Error

In this section, we introduce a bound on the generalization risk between a given function $G_1 \in \mathcal{H}$ and an unknown target function $y_{AB}$, i.e., $R_{D_A}[G_1, y_{AB}]$. This bound is based on a bias-variance decomposition and sums two terms: the bias error and the approximation error. The bias error is the maximal risk possible with a member $G_2$ of the class $\mathcal{P}(\mathcal{H}; \epsilon_1)$, i.e., $\sup_{G_2 \in \mathcal{P}(\mathcal{H}; \epsilon_1)} R_{D_A}[G_1, G_2]$. The approximation error is the minimal possible risk between a member $G$ of the class $\mathcal{P}(\mathcal{H}; \epsilon_1)$ with respect to $y_{AB}$, i.e., $\inf_{G \in \mathcal{P}(\mathcal{H}; \epsilon_1)} R_{D_A}[G, y_{AB}]$.

### 3.1 Derivation of the Bound and the Algorithms

The bound is a consequence of using a loss $\ell$ that satisfies the triangle inequality. Losses of this type include the $L_1$ loss, which is often used in cross domain mapping.

The $L_2$ loss satisfies the triangle inequality up to a factor of three, which would incur the addition of a factor into the bound.

**Lemma 2.** *Let $A = (\mathcal{X}_A, D_A)$ and $B = (\mathcal{X}_B, D_B)$ be two domains, $\mathcal{U} = \{\mathcal{H}_i\}_{i \in I}$ be a family of hypothesis classes and $\epsilon_1 > 0$. In addition, assume that $\ell$ is a loss function that satisfies the triangle inequality. Then, for all $\mathcal{H} \in \mathcal{U}$ such that $\mathcal{P}(\mathcal{H}; \epsilon_1) \neq \emptyset$ and two functions $y_{AB}$ and $G_1$, we have:*

$$R_{D_A}[G_1, y_{AB}] \leq \sup_{G_2 \in \mathcal{P}(\mathcal{H}; \epsilon_1)} R_{D_A}[G_1, G_2] + \inf_{G \in \mathcal{P}(\mathcal{H}; \epsilon_1)} R_{D_A}[G, y_{AB}] \quad (7)$$

**Proof** Let $G^* = \arg\inf_{G \in \mathcal{P}(\mathcal{H}; \epsilon_1)} R_{D_A}[G, y_{AB}]$. By the triangle inequality, we have:

$$\begin{aligned} R_{D_A}[G_1, y_{AB}] &\leq R_{D_A}[G_1, G^*] + R_{D_A}[G^*, y_{AB}] \\ &\leq \sup_{G_2 \in \mathcal{P}(\mathcal{H}; \epsilon_1)} R_{D_A}[G_1, G_2] + \inf_{G \in \mathcal{P}(\mathcal{H}; \epsilon_1)} R_{D_A}[G, y_{AB}] \end{aligned} \quad (8)$$

□ ∎

If $y_{AB}$ satisfies Occam's razor, then the approximation error is lower than $\epsilon_2$ and by Eq. 7 in Lem. 2 the following bound is obtained:

$$R_{D_A}[G_1, y_{AB}] \leq \sup_{G_2 \in \mathcal{P}(\mathcal{H}; \epsilon_1)} R_{D_A}[G_1, G_2] + \epsilon_2 \quad (9)$$

Eq. 9 provides us with an accessible bound for the generalization risk. The right hand side can be directly approximated by training a neural network $G_2$ that has a discrepancy lower than $\epsilon_1$ and has the maximal risk with regards to $G_1$, i.e.,

$$\sup_{G_2 \in \mathcal{H}} R_{D_A}[G_1, G_2] \text{ s.t: } \text{disc}(G_2 \circ D_A, D_B) \leq \epsilon_1 \quad (10)$$

By applying Lagrange relaxation, we obtain the following Lagrangian dual form:

$$L(G_2, \lambda) = R_{D_A}[G_1, G_2] + \mu \cdot (\epsilon_1 - \text{disc}(G_2 \circ D_A, D_B)) \quad (11)$$

Therefore, instead of computing Eq. 10, we maximize the dual form in Eq. 11. For convenience, we will use the following equivalent representation of it:

$$\begin{aligned} &\max_{G_2} R_{D_A}[G_1, G_2] + \mu \cdot (\epsilon_1 - \text{disc}(G_2 \circ D_A, D_B)) \\ \iff &\min_{G_2} -R_{D_A}[G_1, G_2] - \mu \cdot (\epsilon_1 - \text{disc}(G_2 \circ D_A, D_B)) \\ \iff &\min_{G_2} \text{disc}(G_2 \circ D_A, D_B) - (1/\mu) \cdot R_{D_A}[G_1, G_2] - \epsilon_1 \\ \stackrel{\lambda := 1/\mu}{\iff} &\min_{G_2} \text{disc}(G_2 \circ D_A, D_B) - \lambda R_{D_A}[G_1, G_2] \end{aligned} \quad (12)$$

The expectation over $x \sim D_A$ (resp $x \sim D_B$) in the risk and discrepancy are replaced, as is often done, with the sum over the training samples in domain $A$ (resp $B$). Based on this, we present a stopping criterion in Alg. 1, and a method for hyperparameter selection in Alg. 2. Eq. 11 is manifested in Step 4 of the former and Step 6 of the latter is the selection criterion that appears as the last line of both algorithms.

**Algorithm 1** Deciding when to stop training $G_1$
---
**Require:** $S_A$ and $S_B$: unlabeled training sets; $\mathcal{H}$: a hypothesis class; $\epsilon_1$: a threshold; $\lambda$: a trade-off parameter; $T_2$: a fixed number of epochs for $G_2$; $T_1$: a maximal number of epochs.
1: Initialize $G_1^0 \in \mathcal{H}$ and $G_2^0 \in \mathcal{H}$ randomly.
2: **for** $i = 1, ..., T_1$ **do**
3:     Train $G_1^{i-1}$ for one epoch to minimize $\text{disc}(G_1^{i-1} \circ D_A, D_B)$, obtaining $G_1^i$.
4:     Train $G_2^i$ for $T_2$ epochs to minimize $\text{disc}(G_2^i \circ D_A, D_B) - \lambda R_{D_A}[G_1^i, G_2^i]$.
    ▷ $T_2$ provides a fixed comparison point.
5: **end for**
6: **return** $G_1^t$ such that: $t = \arg\min_{i \in [T]} R_{D_A}[G_1^i, G_2^i]$.

---

**Algorithm 2** Model Selection
---
**Require:** $S_A$ and $S_B$: unlabeled training sets; $\mathcal{U} = \{\mathcal{H}_i\}_{i \in I}$: a family of hypothesis classes; $\epsilon$: a threshold; $\lambda$: a trade-off parameter.
1: Initialize $J = \emptyset$.
2: **for** $i \in I$ **do**
3:     Train $G_1^i \in \mathcal{H}_i$ to minimize $\text{disc}(G_1^i \circ D_A, D_B)$.
4:     **if** $\text{disc}(G_1^i \circ D_A, D_B) \leq \epsilon$ **then**
5:         Add $i$ to $J$.
6:         Train $G_2^i \in \mathcal{H}_i$ to minimize $\text{disc}(G_2^i \circ D_A, D_B) - \lambda R_{D_A}[G_1^i, G_2^i]$.
7:     **end if**
8: **end for**
9: **return** $G_1^i$ such that: $i = \arg\min_{j \in J} R_{D_A}[G_1^j, G_2^j]$.

---

### 3.2 Bound on the Loss of Each Sample

We next extend the bound to estimate the error $\ell(G_1(x), y_{AB}(x))$ of mapping by $G_1$ a specific sample $x \sim D_A$. Lem. 3 follows very closely to Lem. 2. It gives rise to a simple method for bounding the loss of $G_1$ on a specific sample $x$. Note that the second term in the bound does not depend on $G_1$ and is expected to be small, since it denotes the capability of overfitting on a single sample $x$.

**Lemma 3.** *Let $A = (\mathcal{X}_A, D_A)$ and $B = (\mathcal{X}_B, D_B)$ be two domains and $\mathcal{H}$ a hypothesis class. In addition, let $\ell$ be a loss function satisfying the triangle inequality. Then, for any target function $y_{AB}$ and $G_1 \in \mathcal{H}$, we have:*

$$\ell(G_1(x), y_{AB}(x)) \leq \sup_{G_2 \in \mathcal{P}(\mathcal{H};\epsilon)} \ell(G_1(x), G_2(x)) + \inf_{G \in \mathcal{P}(\mathcal{H};\epsilon)} \ell(G(x), y_{AB}(x)) \qquad (13)$$

Similarly to the analysis done in Sec. 3, Eq. 13 provides us with an accessible bound for the generalization risk. The RHS can be directly approximated by training

---
**Algorithm 3** Bounding the loss of $G_1$ on sample $x$
---
**Require:** $S_A$ and $S_B$: unlabeled training sets; $\mathcal{H}$: a hypothesis class; $G_1 \in \mathcal{H}$: a mapping; $\lambda$: a trade-off parameter; $x$: a specific sample.
 1: Train $G_2 \in \mathcal{H}$ to minimize $\text{disc}(G_2 \circ D_A, D_B) - \lambda \ell(G_1(x), G_2(x))$.
 2: **return** $\ell(G_1(x), G_2(x))$.
---

a neural network $G_2$ of a discrepancy lower than $\epsilon$ and has maximal loss with regards to $G_1$, i.e.,

$$\sup_{G_2 \in \mathcal{H}} \ell(G_1(x), G_2(x)) \text{ s.t: } \text{disc}(G_2 \circ D_A, D_B) \leq \epsilon \tag{14}$$

With similar considerations as in Sec. 3, we replace Eq. 14 with the following objective:

$$\min_{G_2 \in \mathcal{H}} \text{disc}(G_2 \circ D_A, D_B) - \lambda \ell(G_1(x), G_2(x)) \tag{15}$$

As before, the expectation over $x \sim D_A$ and $x \sim D_B$ in the discrepancy are replaced with the sum over the training samples in domain $A$ and $B$ (resp.).

In practice, we modify Eq. 15 such that $x$ is weighted to half the weight of all samples, during the training of $G_2$. This emphasizes the role of $x$ and allows us to train $G_2$ for less epochs. This is important, as a different $G_2$ must be trained for measuring the error of each sample $x$.

### 3.3 Deriving an Unsupervised Variant of Hyperband using the Bound

In order to optimize multiple hyperparameters simultaneously, we create an unsupervised variant of the hyperband method [4]. Hyperband requires the evaluation of the loss for every configuration of hyperparameters. In our case, our loss is the risk function, $R_{D_A}[G_1, y_{AB}]$. Since we cannot compute the actual risk, we replace it with our bound $\sup_{G_2 \in \mathcal{P}(\mathcal{H};\epsilon_1)} R_{D_A}[G_1, G_2]$.

In particular, the function 'run_then_return_val_loss' in the hyperband algorithm (Alg. 1 of [4]), which is a plug-in function for loss evaluation, is provided with our bound from Eq. 9 after training $G_2$, as in Eq. 11. Our variant of this function is listed in Alg. 4. It employs two additional procedures that are used to store the learned models $G_1$ and $G_2$ at a certain point in the training process and to retrieve these to continue the training for a set amount of epochs. The retrieval function is simply a map between a vector of hypermarkets and a tuple of the learned networks and the number of epochs $T$ when stored. For a new vector of hyperparameters, it returns $T = 0$ and two randomly initialized networks, with architectures that are determined by the given set of hyperparameters. When a network is retrieved, it is then trained for a number of epochs that is the difference between the required number of epochs $T$, which is given by the hyperband method, and the number of epochs it was already trained, denoted by $T_{\text{last}}$.

**Algorithm 4** Unsupervised run_then_return_val_loss for hyperband
───────────────────────────────────────────────
**Require:** $S_A$, $S_B$, and $\lambda$ as before. $T$: Number of epochs. $\theta$: Set of hyperparameters
 1: $[G_1, G_2, T_{\text{last}}]$ = return_stored_functions($\theta$)
 2: Train $G_1$ for $T - T_{\text{last}}$ epochs to minimize $\text{disc}(G_1 \circ D_A, D_B)$.
 3: Train $G_2$ for $T - T_{\text{last}}$ epochs to minimize $\text{disc}(G_2 \circ D_A, D_B) - \lambda R_{D_A}[G_1, G_2]$.
 4: store_functions($\theta$, $[G_1, G_2, T]$)
 5: **return** $R_{D_A}[G_1, G_2]$.
───────────────────────────────────────────────

Table 1: Pearson correlations and the corresponding p-values (in parentheses) of the ground truth error with: (i) the bound, (ii) the GAN losses, and (iii) the circularity losses or (iv) the distance correlation loss. *The cycle loss $A \to B \to A$ is shown for DiscoGAN and the distance correlation loss is shown for DistanceGAN.

| Alg. | Method | Dataset | Bound | $GAN_A$ | $GAN_B$ | $Cycle_A/\mathcal{L}_D^*$ | $Cycle_B$ |
|---|---|---|---|---|---|---|---|
| Alg. 1 | Disco-GAN [8] | Shoes2Edges | **1.00** (<1E-16) | -0.15 (3E-03) | -0.28 (1E-08) | 0.76 (<1E-16) | 0.79 (<1E-16) |
| | | Bags2Edges | **1.00** (<1E-16) | -0.26 (6E-11) | -0.57 (<1E-16) | 0.85 (<1E-16) | 0.84 (<1E-16) |
| | | Cityscapes | **0.94** (<1E-16) | -0.66 (<1E-16) | -0.69 (<1E-16) | -0.26 (1E-07) | 0.80 (<1E-16) |
| | | Facades | **0.85** (<1E-16) | -0.46 (<1E-16) | 0.66 (<1E-16) | 0.92 (<1E-16) | 0.66 (<1E-16) |
| | | Maps | **1.00** (<1E-16) | -0.81 (<1E-16) | 0.58 (<1E-16) | 0.20 (9E-05) | -0.14 (5E-03) |
| | Distance-GAN [14] | Shoes2Edges | **0.98** (<1E-16) | - | -0.25 (2E-16) | -0.14 (1E-05) | - |
| | | Bags2Edges | **0.93** (<1E-16) | - | -0.08 (2E-02) | 0.34 (<1E-16) | - |
| | | Cityscapes | **0.59** (<1E-16) | - | 0.22 (1E-11) | -0.41 (<1E-16) | - |
| | | Facades | **0.48** (<1E-16) | - | 0.03 (5E-01) | -0.01 (9E-01) | - |
| | | Maps | **1.00** (<1E-16) | - | -0.73 (<1E-16) | 0.39 (4E-16) | - |
| Alg. 2 | Disco-GAN [8] | Shoes2Edges | **0.95** (1E-03) | 0.73 (7E-02) | 0.51 (2E-01) | 0.05 (9E-01) | 0.05 (9E-01) |
| | | Bags2Edges | **0.99** (2E-06) | 0.64 (2E-01) | 0.54 (3E-01) | -0.26 (7E-01) | -0.20 (7E-01) |
| | | Cityscapes | **0.99** (1E-03) | 0.69 (9E-02) | 0.85 (2E-02) | -0.53 (2E-01) | -0.42 (4E-01) |
| | | Facades | **0.94** (1E-03) | -0.33 (4E-01) | 0.88 (4E-02) | 0.66 (8E-02) | -0.45 (3E-01) |
| | | Maps | **1.00** (1E-03) | 0.62 (1E-01) | 0.54 (2E-01) | 0.60 (2E-01) | 0.07 (9E-01) |
| | Distance-GAN [14] | Shoes2Edges | **0.96** (1E-04) | - | 0.33 (5E-01) | -0.87 (6E-03) | - |
| | | Bags2Edges | **0.98** (1E-05) | - | -0.11 (8E-01) | 0.23 (6E-01) | - |
| | | Cityscapes | **0.92** (1E-03) | - | 0.66 (8E-02) | -0.49 (2E-01) | - |
| | | Facades | **0.84** (2E-02) | - | 0.75 (5E-02) | 0.37 (4E-01) | - |
| | | Maps | **0.95** (1E-03) | - | -0.43 (3E-01) | -0.15 (7E-01) | - |
| Alg. 3 | Disco-GAN [8] | Shoes2Edges | **0.92** (<1E-16) | -0.12 (5E-01) | 0.02 (9E-01) | 0.29 (6E-02) | 0.15 (4E-01) |
| | | Bags2Edges | **0.96** (<1E-16) | 0.25 (1E-01) | 0.08 (6E-01) | 0.08 (6E-01) | 0.05 (7E-01) |
| | | Cityscapes | **0.78** (4E-04) | 0.24 (4E-01) | -0.16 (6E-01) | -0.04 (9E-01) | 0.03 (9E-01) |
| | | Facades | **0.80** (6E-10) | 0.13 (4E-01) | 0.16 (3E-01) | 0.20 (2E-01) | 0.09 (5E-01) |
| | | Maps | **0.66** (1E-03) | 0.08 (7E-01) | 0.12 (6E-01) | 0.17 (5E-01) | -0.25 (3E-01) |
| | Distance-GAN [14] | Shoes2Edges | **0.98** (<1E-16) | - | -0.05 (7E-01) | 0.84 (<1E-16) | - |
| | | Bags2Edges | **0.92** (<1E-16) | - | -0.28 (2E-01) | 0.45 (3E-02) | - |
| | | Cityscapes | **0.51** (4E-04) | - | 0.10 (5E-01) | 0.28 (2E-2) | - |
| | | Facades | **0.72** (<1E-16) | - | -0.01 (1E00) | 0.08 (6E-01) | - |
| | | Maps | **0.94** (1E-06) | - | 0.20 (2E-01) | 0.30 (6E-02) | - |

## 4. Experiments

We test the three algorithms on two unsupervised alignment methods: DiscoGAN [8] and DistanceGAN [14]. In DiscoGAN, we train $G_1$ (and $G_2$), using two GANs and two circularity constraints; in DistanceGAN, one GAN and one distance correlation loss are used. The published parameters for each dataset are used, except when applying our model selection method, where we vary the number of layers and when using hyperband, where we vary the learning rate and the batch size as well.

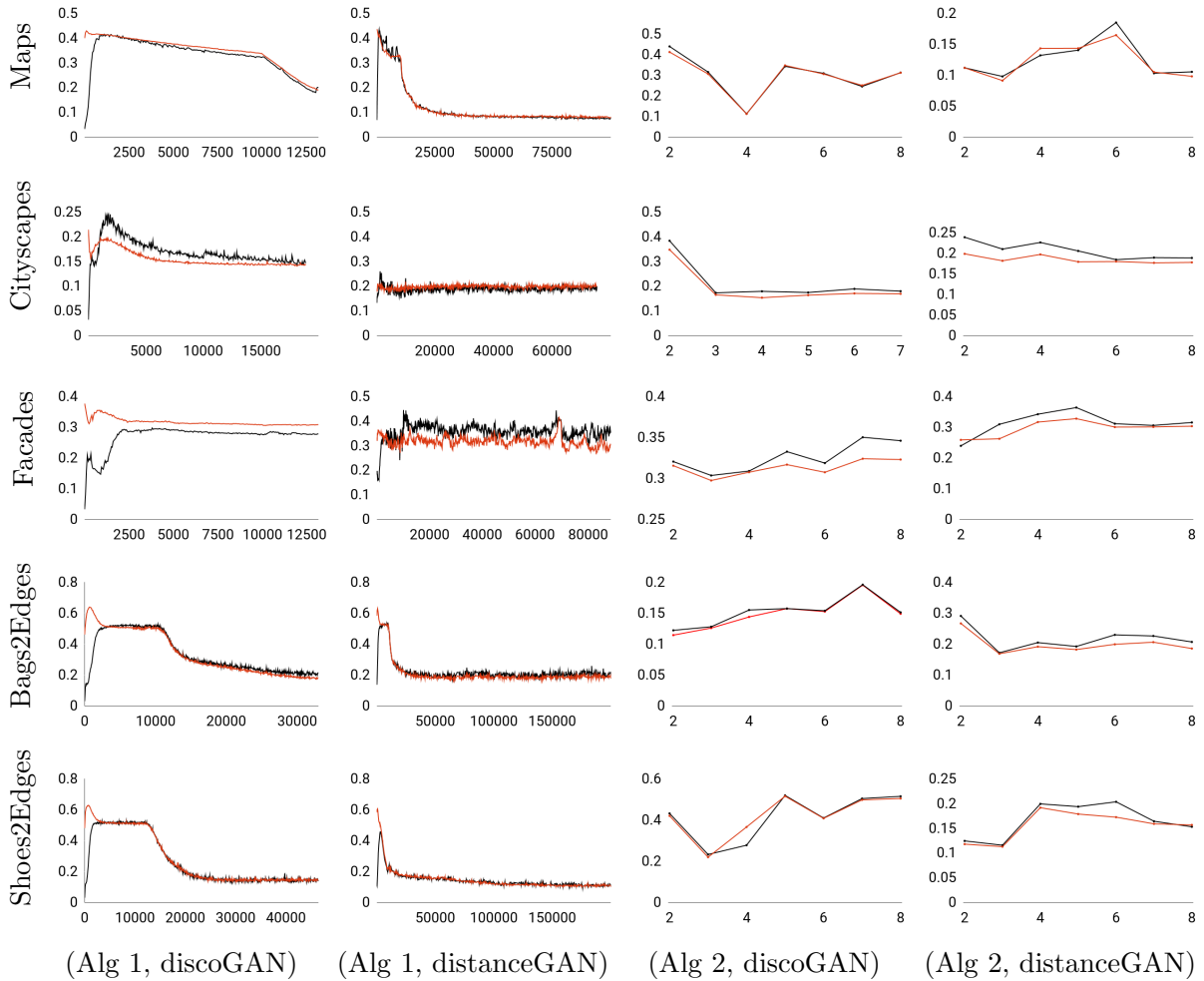

Figure 2: Results of Alg. 1, 2. Ground truth errors are in red and bound in black. x-axis is the iteration or number of layers. y-axis is expected risk. For Alg. 1 it takes a few epochs for $G_1$ to have a small enough discrepancy, until which the bound is ineffective.

Five datasets were used in the experiments: (i) aerial photographs to maps, trained on data scraped from Google Maps [13], (ii) the mapping between photographs from the cityscapes dataset and their per-pixel semantic labels [25], (iii) architectural photographs to their labels from the CMP Facades dataset [26], (iv) handbag images [27] to their binary edge images as obtained from the HED edge detector [28], and (v) a similar dataset for the shoe images from [29].

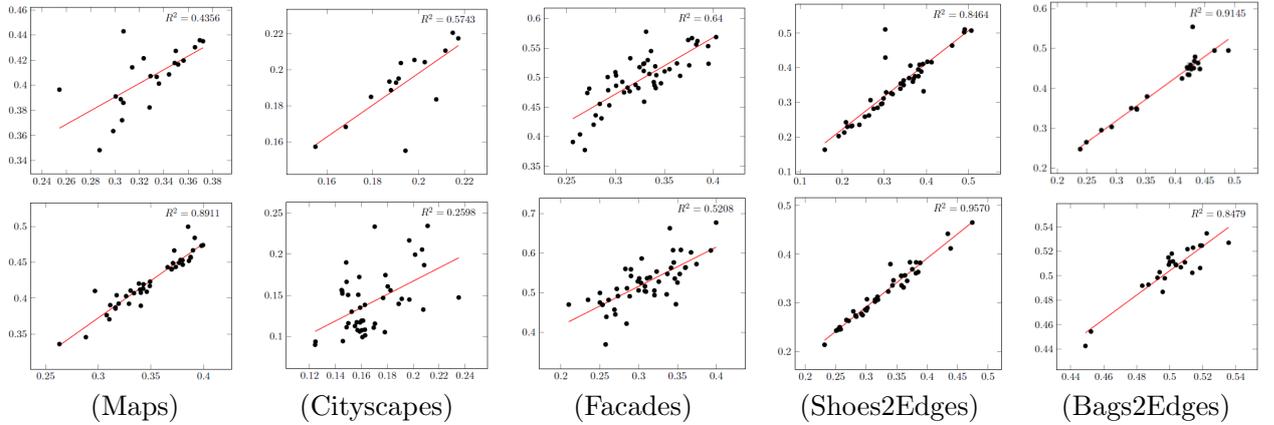

Figure 3: Results of Alg. 3. Results shown for DiscoGAN for the first row and for DistanceGAN in the second row. The ground truth errors (x-axis) vs. bound (y-axis) are shown per point. The coefficient of determination is shown (top right).

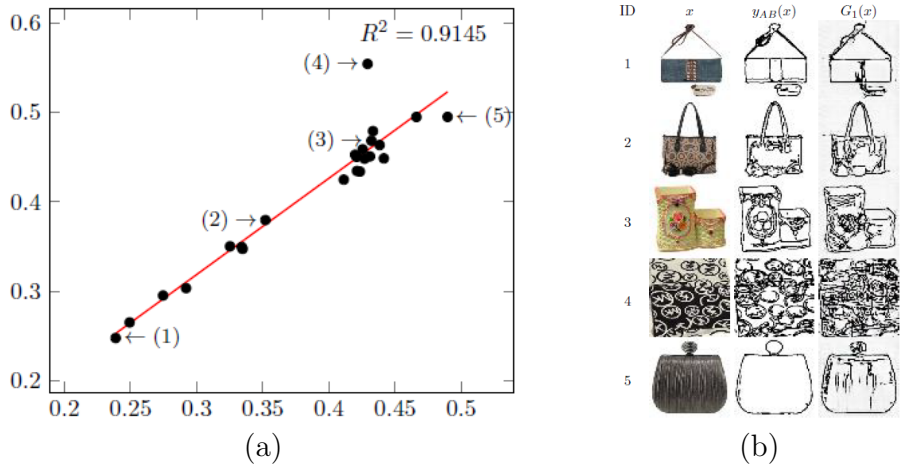

Figure 4: Results of Alg. 3 on DiscoGAN bags2edges. (a) The ground truth errors vs. the bound per point are shown. This is the same as Fig. 3 top right plot with added information identifying specific points. (b) The source (x), ground truth ($y_{AB}(x)$) and mapping ($G_1(x)$) of the marked points.

Throughout the experiments, fixed values are used as the low-discrepancy threshold ($\epsilon_1 = 0.2$). The tradeoff parameter between the dissimilarity term and the fitting term during the training of $G_2$ is set, per dataset, to be the maximal value such that the fitting of $G_2$ provides a solution that has a discrepancy lower than the threshold, $\text{disc}(G_2 \circ D_A, D_B) \leq \epsilon_1$. This is done once, for the default parameters of $G_1$, as given in the original DiscoGAN and DistanceGAN [8, 14].

The results of all experiments are summarized in Tab. 1, which presents the correlation and p-value between the ground truth error, as a function of the independent

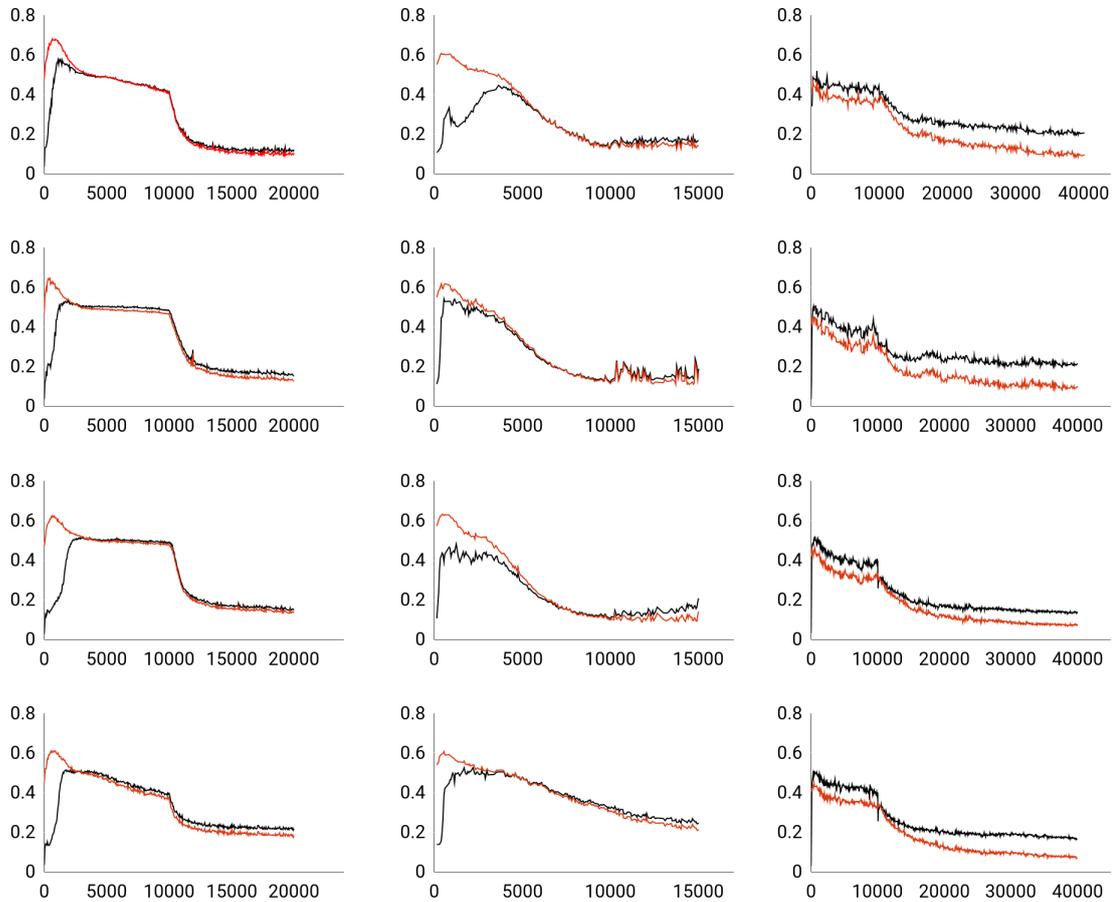

(Handbags2Edges, discoGAN)    (Shoes2Edges, distanceGAN)    (Maps, distanceGAN)

Figure 5: Per-epoch per-sample results for three experiments, four points each. x-axis is iteration. y-axis is the per-sample error. Red line indicates the ground truth error of an individual sample, i.e $||G_1(x) - y(x)||_1$. Black line indicates our bound for an individual sample, i.e $||G_1(x) - G_2(x)||_1$. Note that it takes a few epochs for $G_1$ to have a small enough discrepancy, until which the bound is ineffective.

variable, and the bound. The independent variable is either the training epoch, the architecture, or the sample, depending on the algorithm tested. For example, in Alg. 2 we wish to decide on the best architecture, the independent variable is the number of layers. A high correlation (low p-value) between the bound and the ground truth error, both as a function of the number of layers, indicates the validity of the bound and the utility of the algorithm. Similar correlations are shown with the GAN losses and the reconstruction losses (DiscoGAN) or the distance correlation loss (Distance-GAN), in order to demonstrate that these are much less correlated with the ground truth error. In the plots of Fig. 2, we omit the other scores in order to reduce clutter.

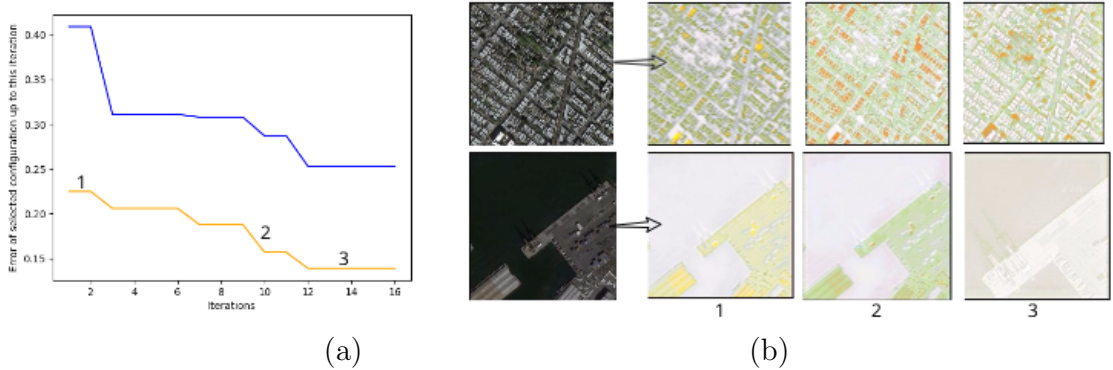

(a)             (b)

Figure 6: Applying unsupervised hyperband for selecting the best configuration for UNIT for the Maps dataset. (a) blue and orange lines are bound and ground truth error as in Fig. 7. (b) Images produced for 3 different configurations as indicated on the plot in (a).

**Stopping Criterion (Alg. 1)** For testing the stopping criterion suggested in Alg. 1, we compared, at each time point, two scores that are averaged over all training samples: $||G_1(x) - G_2(x)||_1$, which is our bound, and the ground truth error $||G_1(x) - y_{AB}(x)||_1$, where $y_{AB}(x)$ is the ground truth image that matches $x$ in domain $B$.

Note that similar to the experiments with ground truth in the literature [8, 9, 14], the ground truth error is measured in the label space and not in the image domain. The mapping in the other direction $y_{BA}$ is not one to one.

The results are depicted in the main results table (Tab. 1) as well as in Fig. 2 for both DiscoGAN (first column) and DistanceGAN (second column). As can be seen, there is an excellent match between the mean ground truth error of the learned mapping $G_1$ and the predicted error. No such level of correlation is present when considering the GAN losses or the reconstruction losses (for DiscoGAN), or the distance correlation loss of DistanceGAN. Specifically, the very low p-values in the first column of Tab. 1 show that there is a clear correlation between the ground truth error and our bound for all datasets. For the other columns, the values in question are chosen to be the losses used for $G_1$. The lower scores in these columns show that none of these values are as correlated with the ground truth error, and so cannot be used to estimate this error.

In the experiment of Alg. 1 for DiscoGAN, which has a large number of sample points, the cycle from $B$ to $A$ and back to $B$ is significantly correlated with the ground truth error with very low p-values in four out of five datasets. However, its correlation is significantly lower than that of our bound.

In Fig. 2, the Facades graph shows a different behavior than the other graphs. This is because the Facades dataset is inherently ambiguous and presents multiple possible mappings from $A$ to $B$. Each mapping satisfies the Occam's razor property separately.

**Selecting Architecture using Alg. 2** Next we vary the number of layers of $G$ and consider its effect on the risk by measuring the bound and the ground truth error (which cannot be computed in an unsupervised way); A large correlation between our bound and the ground truth error is observed, see Tab. 1 and Fig. 2, columns 3 and 4. We can therefore optimize the number of layers based on our bound. With a much smaller number of sample points, the p-values are generally higher than in the previous experiment.

Beyond correlations, Fig 2 (all four columns), can be used to quantify the gain from using the two algorithms. The "regret" when using the algorithm is simply the ground truth error at the minimal value of the bound minus the minimal ground truth error.

**Predicting per-Sample Loss with Alg. 3** Finally, we consider the per sample loss. The results are reported numerically in Tab. 1 and plotted in Fig. 3, 4. As can be seen, there is a high degree of correlation between the measured bound and the ground truth error. Therefore, our method is able to reliably predict the per-sample success of a multivariate mapping learned in a fully unsupervised manner.

Remarkably, this correlation also seems to hold when considering the time axis, i.e., we can combine Alg. 1 and Alg. 3 and select the stopping epoch that is best for a specific sample. Fig. 5 depicts, for three experiments, the bound and the per-sample loss of $G_1$ over time. In each graph, we plotted the values of the bound and the loss over time during training of $G_1$. In each column we have the results for four samples with a specific dataset and method. As can be seen, in the datasets tested, the bound holds over time. However, the points of a specific dataset seem to follow relatively similar patterns of improvement in time.

**Selecting Architecture with the Modified Hyperband Algorithm** Our bound is used in Sec. 3.3 to create an unsupervised variant of the hyperband method. In comparison to Alg. 2, this allows for the optimization of multiple hyperparameters at once, while enjoying the efficient search strategy of the hyperband method.

Fig. 7 demonstrates the applicability of our unsupervised hyperband-based method for different datasets, employing both DiscoGAN and DistanceGAN. The graphs show the error and the bound obtained for the selected configuration after up to 35 hyperband iterations. As can be seen, in all cases, the method is able to recover a configuration that is significantly better than what is recovered, when only optimizing for the number of layers. To further demonstrate the generality of our method, we applied it on the UNIT [30] architecture. As the runtime of UNIT is much higher than DiscoGAN and DistanceGAN, this did not allow for extensive experimentation. We therefore focused on the most useful application of applying hyperband on a relatively complex dataset, specifically Maps. Fig. 6 and Tab. 7(b) show the convergence on the hyperband method.

## 5. Conclusions

We extend the envelope of what is known to be possible in unsupervised learning by showing that we can reliably predict the error of a cross-domain mapping that was trained without matching samples. This is true both in expectation, with application to hyperparameter selection, and per sample, thus supporting dynamic confidence-based run time behavior, and (future work) unsupervised boosting during training.

The method is based on measuring the maximal distance within the set of low discrepancy mappings. This measure becomes the bound by applying what we define as the Occam's razor property, which is a general form of the Simplicity Principle. Therefore, the clear empirical success observed in our experiments supports the recent hypothesis that simplicity plays a key role in unsupervised learning.

## Acknowledgements

This project has received funding from the European Research Council (ERC) under the European Union's Horizon 2020 research and innovation programme (grant ERC CoG 725974).

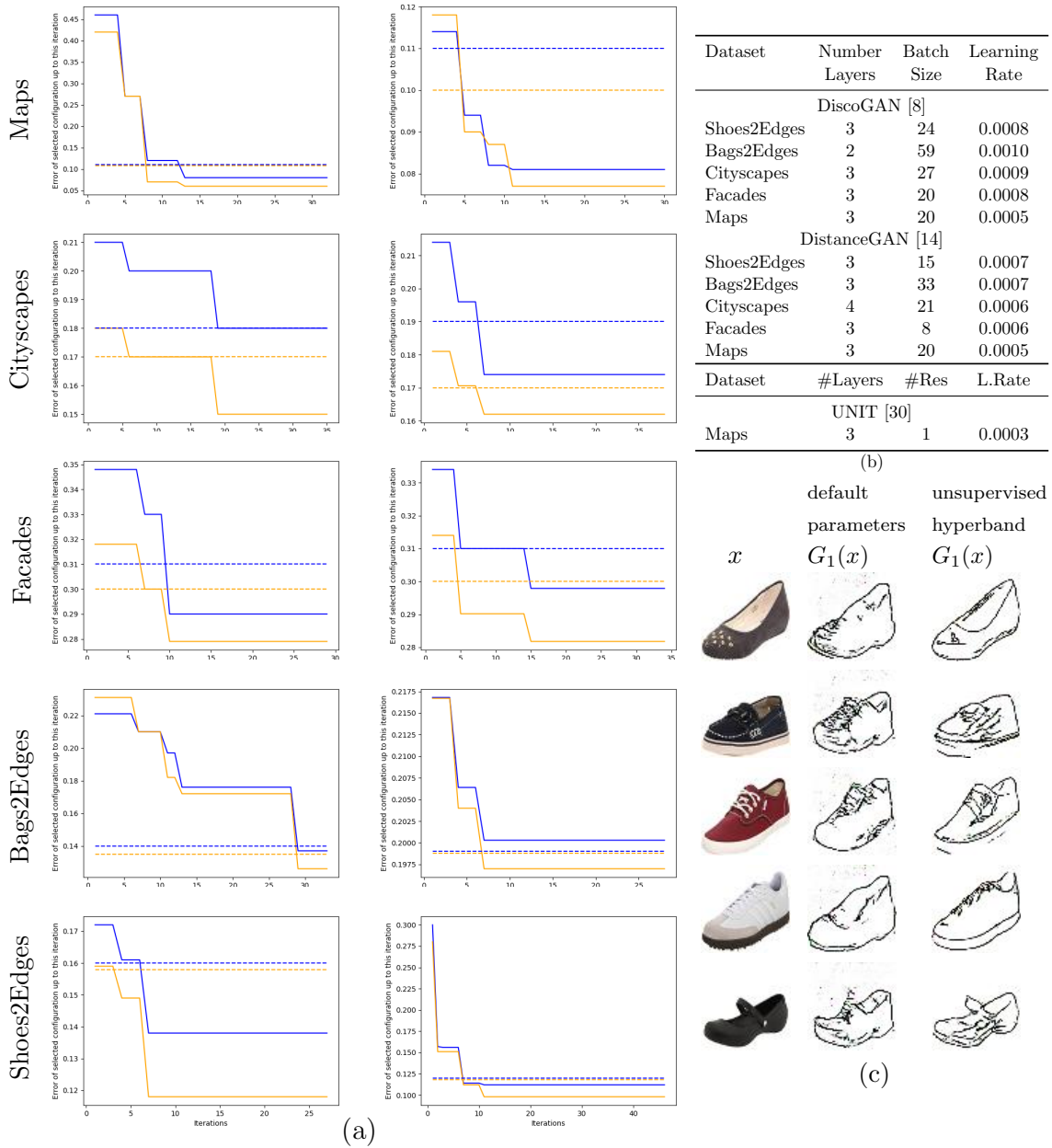

Figure 7: Applying unsupervised hyperband for selecting the best configuration. For DiscoGAN and DistanceGAN we optimize of the number of encoder and decoder layers, batch size and learning rate while for UNIT, we optmized for the number of encoder and decoder Layers, number of resnet layers and learning rate. (a) For each dataset, the first plot is of DiscoGAN and the second is of DistanceGAN. Hyperband optimizes according to the bound values indicated in blue. The corresponding ground truth errors are shown in orange. Dotted lines represent the best configuration errors, when varying only the number of layers without hyperband (blue for bound and orange for ground truth error). Each graph shows the error of the best configuration selected by hyperband as a function the number of hyperband iterations. (b) The corresponding hyperparameters of the best configuration as selected by hyperband. (c) Images produced for DiscoGAN's shoes2edges: 1st column is the input, the 2nd is the result of DiscoGAN's default configuration, 3rd is the result of the configuration selected by our unsupervised Hyperband.